\documentclass[10pt]{article}
\usepackage{fancyhdr}
\usepackage{extramarks}
\usepackage{amsmath}
\usepackage{amsthm}
\usepackage{amsfonts}
\usepackage{siunitx}
\usepackage{tikz}
\usepackage[plain]{algorithm}
\usepackage{algpseudocode}
\usepackage{multirow}
\usepackage{booktabs}
\usepackage{graphicx}
\usepackage{subfigure}
\usepackage[colorlinks,linkcolor=black,anchorcolor=black,citecolor=black,urlcolor=blue]{hyperref}
\usepackage{amsmath,bm}
\usepackage{booktabs}
\usepackage{mathtools}
\usepackage{amssymb}
\usepackage{caption}
\usepackage{capt-of}
\usepackage{mciteplus}
\usepackage{cite}
\usepackage{mathrsfs}
\usepackage[title,titletoc,toc]{appendix}
\usepackage{xr}
\usepackage{parskip}
\usetikzlibrary{automata,positioning}
\usepackage{wrapfig}
\renewcommand{\part}[1]{\textbf{\large Part \Alph{partCounter}}\stepcounter{partCounter}\\}

\graphicspath{{./figures/}}
\begin{document}

\title{Analyzing scRNA-seq data by CCP-assisted UMAP and t-SNE }
\author{Yuto Hozumi$^1$ and Guo-Wei Wei$^{1,2,3}$\footnote{
			Corresponding author.		Email: weig@msu.edu} \\
		$^1$ Department of Mathematics, \\
		Michigan State University, MI 48824, USA.\\
		$^2$ Department of Electrical and Computer Engineering,\\
		Michigan State University, MI 48824, USA. \\
		$^3$ Department of Biochemistry and Molecular Biology,\\
		Michigan State University, MI 48824, USA. \\
	}
\maketitle

\begin{abstract} 
 
Single-cell RNA sequencing (scRNA-seq) is widely used to reveal heterogeneity in cells, which has given us insights into cell-cell communication, cell differentiation, and differential gene expression. However, analyzing scRNA-seq data is a challenge due to sparsity and the large number of genes involved. Therefore, dimensionality reduction and feature selection are important for removing spurious signals and enhancing downstream analysis.
Correlated clustering and projection (CCP) was recently introduced as an effective method for preprocessing scRNA-seq data. CCP utilizes gene-gene correlations to partition the genes and, based on the partition,  employs cell-cell interactions to obtain super-genes. Because CCP is a data-domain approach that does not require matrix diagonalization, it can be used in many downstream machine learning tasks. In this work, we utilize CCP as an initialization tool for uniform manifold approximation and projection (UMAP) and t-distributed stochastic neighbor embedding (t-SNE). By using eight publicly available datasets, we have found that CCP significantly  improves UMAP and t-SNE visualization and dramatically improve their accuracy.

\end{abstract}

keywords: {scRNA-seq, dimensionality reduction, machine learning, visualization}

\newpage

\section{Introduction}

Single-cell RNA sequencing (scRNA-seq) is a relatively new technology that profiles the transcriptome of individual cells within a tissue or organ, aiming to gain understanding of gene expression, gene regulation, cell-cell interaction, spatial transcriptomics, signal transduction pathways, and more \cite{lun2016step}. The typical workflow of scRNA-seq involves cell isolation, RNA extraction, library preparation, sequencing, and data analysis.
One of the key challenges in scRNA-seq analysis is the enormous amount of data generated, which can be complex, nonuniform, noisy, unlabeled, and of excessively high dimensions. A typical data analysis pipeline involves data preprocessing, gene expression quantification, normalization and batch correction, dimensionality reduction, cell type identification, differential gene expression analysis, and pathway and functional analysis \cite{hwang2018single, andrews2021tutorial, luecken2019current, chen2019single, petegrosso2020machine, li2019statistical}. Principal component analysis (PCA), t-distributed stochastic neighbor embedding (t-SNE), and uniform manifold approximation and projection (UMAP) are some of the most popular approaches for dimensionality reduction, clustering, and visualization of scRNA-seq data.

PCA is a classical  technique used for   dimensionality reduction and visualization. It identifies the most important patterns and correlations in a high-dimensional dataset and expresses them as a linear combination of new and orthogonal components. Among them, the first few components are often regarded as principal components, which can be used to visualize the scRNA-seq data in a lower-dimensional space or to identify important gene expression patterns. PCA remains a key technique in scRNA-seq \cite{zhou2022pca}.

The t-SNE algorithm is a nonlinear statistical learning approach that helps visualize high-dimensional scRNA-seq data in a lower-dimensional space of two or three dimensions while preserving the pairwise similarities between cell samples \cite{van2008visualizing,kobak2019art}. It maps high-dimensional data into a lower-dimensional space by minimizing the difference between the high-dimensional and low-dimensional probability distributions via the Kullback–Leibler divergence (KL divergence) using a gradient descent algorithm. T-SNE has been widely used in scRNA-seq analysis, particularly for the visualization of cell clustering and gene expression patterns.  
 
UMAP is another machine learning algorithm for scRNA-seq dimensionality reduction, clustering, and visualization \cite{mcinnes2018umap}. It constructs a low-dimensional weighted graph to capture the local connectivity of high-dimensional data. The optimization of graph coefficients through the KL divergence preserves the global structure or distribution of the data in the low-dimensional weighted graph. It is often faster and more scalable than t-SNE and can handle larger datasets. UMAP is one of the most popular visualization tools in computational biology and bioinformatics. UMAP-assisted  $k$-means clustering was developed to deal with large scale datasets \cite{hozumi2021umap}. 

PCA, t-SNE, UMAP, and many other dimensionality reduction methods rely on matrix diagonalization, which comes with a few drawbacks. First, the resulting components are in the space associated with the matrix eigenvectors, which do not have a direct physical interpretation. This gives rise to challenges in the downstream analysis of  outliers. Additionally, matrix diagonalization can be expensive and inaccessible to a large number of samples, i.e., the so-called ``curse of dimensionality''. Moreover, the procedure can be very unstable with respect to data noise, missing data,  low quality data, etc. Therefore, data initialization is  a critical step to preserve the global data structure in both t-SNE and UMAP \cite{kobak2021initialization}. In general, PCA, t-SNE, and UMAP work extremely well for problems with low intrinsic dimensions but may not be efficient and robust for truly high dimensional problems.   

Many non-negative matrix factorization (NMF) methods have been proposed  with different constraints, such that the low-dimensional representation of scRNA-seq is a linear combination of the original data, and can be interpreted as meta-genes \cite{shu2022robust, wu2020robust, chen2022detecting, xiao2018graph, yu2019robust, liu2017joint}. ScRNA by non-negative and low-rank representation (SinLRR) is designed to find the smallest rank matrix that can capture the original data \cite{zheng2019sinnlrr}. A cell-cell similarity metric is learned in  single-cell interpretation via multikernel learning (SIMLR)  to reduce the dimensionality \cite{wang2017visualization}. Various deep learning strategies have been developed  for dimensionality reduction \cite{lopez2018deep, torroja2019digitaldlsorter, yuan2022scmra, luo2021topology, wang2018vasc, lin2020deep}.  Despite of much effort, scRNA seq analysis remains a challenge task, due to   zero expression counts,   low reading counts, sample noise, and the intrinsic high dimensionality associated with tens of thousands of genes \cite{lahnemann2020eleven}.

We proposed correlated clustering and projection (CCP) as a general approach for dimensionality reduction \cite{hozumi2022ccp}. CCP is a data-domain method that completely bypasses matrix diagonalization. It partitions genes into clusters based on their local similarities and then projects genes in the same clusters into a super-gene, which is a measure of accumulated gene-gene correlations among cells. The gene partition can be realized by either the standard $k$-means or the $k$-medoids using either covariance distance or correlation distance. Flexibility rigidity index (FRI) is used for the nonlinear  projection \cite{xia2013multiscale}. The resulting super-genes are all non-negative and highly interpretable. Recently, CCP has been applied to the clustering and classification of scRNA-seq datasets \cite{hozumi2023preprocessing}. Validation   on  fourteen  scRNA-seq datasets indicates that PCA outperforms PCA for problems with a high intrinsic dimensionality.

The objective of the present work is to explore the utility of CCP for initializing scRNA-seq data. We are particularly interested in its potential application for initializing UMAP and t-SNE, which are among the most successful visualization tools in scRNA-seq analysis. We tested CCP-assisted UMAP and CCP-assisted  t-SNE on eight publicly available datasets.
Additionally, we introduce a novel method for handling low-variance (LV) genes. Instead of discarding low-variance genes like many other methods, we group them together into a single category. This grouping is achieved by projecting them into one descriptor using FRI. One of the drawbacks of dropping low-variance genes is that scRNA-seq data often has an unequal number of cell types. Moreover, there are numerous genes with low expression, and removing too many genes may result in overlooking cell outliers. Therefore, LV-gene addresses this issue by consolidating low-variance genes into one descriptor, thereby increasing its predictive power. We found that  CCP  improves the accuracy of UMAP and t-SNE by  over 19\% in each case.

\section{Methods and Algorithms}

Consider a scRNA-seq dataset $\mathcal{Z}  := \{z_{m}^i\}_{m=1, i=1}^{M, I}$ with 
 $M$ and $I$ being the number of input data entries (i.e., samples or cells) and the number of genes, respectively. 
CCP finds an $N$-dimensional representation  
 $\mathcal{X}  := \{x_{m}^n\}_{m=1, n=1}^{M, N}$,  
in which  $1 \le N << I$, by using a data-domain two-step   strategy:   gene clustering  and gene projection. The resulting component $x_{m}^n$ can be regarded as the $n$th super-gene for the $m$th cell.  

\subsection{Gene clustering }
To facilitate a gene clustering, we emphasize   gene vectors by setting the original data as $\mathcal{Z}  = \{\mathbf{z}^1, ..., \mathbf{z}^i, ..., \mathbf{z}^I\}$, where  $\mathbf{z}^i \in \mathbb{R}^M$ represents the $i$th gene vector for the data. CCP  partitions the gene vector into $N$ components with  $1\le N <<I$ by a clustering technique, such as $k$-means or $k$-medoids. CCP seeks   an optimal disjoint partition of the data $\displaystyle \mathcal{Z}  := \uplus_{n=1}^N \mathcal{Z}^{n}$, for a given $N$, where $\mathcal{Z}^{n}$ is the $n$th partition (cluster) of the genes. To this end,  the correlations among gene vectors $  \mathbf{z}^i $ are analyzed according to appropriate  correlation measures, such as   
covariance distance and correlation distance.

	\subsubsection{Gene projection}
Based on the gene partitioning, we denote $\mathbf{z}_m^{S^n}  \in \mathbb{R}^{|S^n|}$ as $n$th cluster of $|S^n|$ genes for the $m$th cell. CCP projects these genes  into a super-gene $x_m^n$ by using the flexibility rigidity index (FRI). Denote $\|\mathbf{z}_i^{S^n} - \mathbf{z}_j^{S^n}\|$ as some metric between  cell  $i$ and  cell $j$ for the $n$th cluster of  $|S^n|$ genes, and the gene-gene correlation between the two cells are defined by $\displaystyle C_{ij}^{S^n} = \Phi(\|\mathbf{z}_i^{S^n} - \mathbf{z}_j^{S^n}\|; \eta^{S^n}, \tau, \kappa)$, 
	where $\Phi$ is a correlation kernel, with parameters $\eta^{S^n}, \tau, $ and $ \kappa > 0 $. One may use   the Euclidean, Manhattan, and/or Wasserstein distances to measure the correlations. Additionally, the FRI correlation kernels satisfy the following   conditions
	\begin{align*} 
		& \Phi(\|\mathbf{z}_i^{S^n} - \mathbf{z}_j^{S^n}\|; \eta^{S^n}, \tau, \kappa) \to 0, \quad \text{as} \|\mathbf{z}_i^{S^n} - \mathbf{z}_j^{S^n}\| \to \infty \\ 
		& \Phi(\|\mathbf{z}_i^{S^n} - \mathbf{z}_j^{S^n}\|; \eta^{S^n}, \tau, \kappa) \to 1, \quad \text{as} \|\mathbf{z}_i^{S^n} - \mathbf{z}_j^{S^n}\| \to 0.
	\end{align*}
	Although various radial basis functions can be used in CCP, we consider generalized exponential function in the present work
	\begin{align*}
		\Phi(\|\mathbf{z}_i^{S^n} - \mathbf{z}_j^{S^n}\|; \eta^{S^n}, \tau, \kappa) = \begin{cases}
			e^{-\left(\frac{\|\mathbf{z}_i^{S^n} - \mathbf{z}_j^{S^n}\|} { \eta^{S^n}\tau}\right)^\kappa} & \|\mathbf{z}_i^{S^n} - \mathbf{z}_j^{S^n}\| < r_c^{S^n} \\
			0, & \text{otherwise}.
		\end{cases}
	\end{align*}
	where $r_c^{S^n}$ is the cutoff distance and $\eta^{S^n}$ is the scale, which are defined from the data automatically. Here, $\kappa$ is the power and $\tau$ is a scale parameter.
	
The gene-gene correlation matrix $C^{S^n} = \{C_{ij}^{S^n}\}$  represents cell-cell interactions.  One  can consider a cutoff $r_c^{S^n}$ to make it sparse. We take the  3-standard deviations of the pairwise distances for $r_c^{S^n}$. To automatically evaluate   $\eta^{S^n}$,  we consider the average minimal distance between the cluster of genes 
	\begin{align*}
		\eta^{S^n} = \frac{\sum_{m=1}^M \min_{\mathbf{z}_j^{S^n}}\|\mathbf{z}_m^{S^n} - \mathbf{z}_j^{S^n}\| }{M}.
	\end{align*}
	Using the correlation function, CCP projects $|S^n|$ genes into a super-gene using FRI for $i$th sample,

\begin{align*}
		x_i^n = \sum_{m=1}^M w_{im}\Phi(\|\mathbf{z}_i^{S^n} - \mathbf{z}_m^{S^n}\|; \eta^{S^n}, \tau, \kappa),
	\end{align*}
where $w_{im}$ are the weights.

CCP obtains the lower dimensional super-gene representation for $i$th sample (cell) $\mathbf{x}_i = (x_i^1, ..., x_i^N)^T$ by running the projection for all gene clusters $ \{\mathcal{Z}^n \}$.  

\subsubsection{Low variance (LV) genes}
Let $\mathbf{v} = (v_1, ..., v_N)$ be the variance of the genes, where $v_i$ is the variance of gene $\mathbf{z}^i$, and assume that the variance are sorted in descending order. Then, define the low variance set $P$ as  
\begin{align*}
	P = \{i | i > v_cN\}
\end{align*}
where $0 \le v_c \le 1$ is the cutoff ratio. Then, we can obtain the cell-cell correlation using these low variance genes $C_{ij}^P$,
\begin{align*}
	C_{ij}^P =  \Phi(\|\mathbf{z}_i^{P} - \mathbf{z}_j^{P}\|; \eta^{P}, \tau, \kappa) 
\end{align*} 
where $\Phi(\|\mathbf{z}_i^{P} - \mathbf{z}_j^{P}\|; \eta^{P}, \tau, \kappa)$ is the generalized exponential function
\begin{align*}
	\Phi(\|\mathbf{z}_i^{P} - \mathbf{z}_j^{P}\|; \eta^{P}, \tau, \kappa) = \begin{cases}
		e^{-\left(\frac{\|\mathbf{z}_i^{P} - \mathbf{z}_j^{P}\|} { \eta^{P}\tau}\right)^\kappa} & \|\mathbf{z}_i^{P} - \mathbf{z}_j^{P}\| < r_c^{P} \\
		0, & \text{otherwise}.
	\end{cases}
\end{align*}
$r_c^P$ is taken as the 3-standard deviation of the pairwise distances, and $\eta^P$ is the average minimum distance 
\begin{align*}
	\eta^{P} = \frac{\sum_{m=1}^M \min_{\mathbf{z}_j^{P}}\|\mathbf{z}_m^{P} - \mathbf{z}_j^{P}\| }{M}.
\end{align*}

Using the correlation function, CCP projects $|P|$ genes into a super-gene using FRI for $i$th sample,

\begin{align*}
	x_i = \sum_{m=1}^M w_{im}\Phi(\|\mathbf{z}_i^{P} - \mathbf{z}_m^{P}\|; \eta^{S^n}, \tau, \kappa),
\end{align*}
where $w_{im}$ are the weights.

For CCP, we compute the LV-gene first, and use the correlated partition algorithm on the remaining genes. 

\section{Results}
\subsection{Data Preprocessing}
We have tested CCP-assisted UMAP and t-SNE visualization on eight publicly available data. \autoref{tab: dataset} shows the accession id from Gene Expression Omnibus \cite{edgar2002gene, barrett2012ncbi}, the reference, source organism, number of samples, number of genes, and the number of cell types.
\begin{table}[H]
			\centering
			\caption{Accession ID, source organism, and the counts for samples, genes, and cell types for fourteen individual datasets}
			\begin{tabular}{|c|c c c c  c c|} \hline
				Accession ID & Reference & Source organism & Samples & Genes & Cell type & Super-genes  \\ \hline
				GSE57249 & Biase \cite{biase2014cell} & Mouse & 49 & 25737 & 3 & 50\\ 
				
				GSE67835 & Darmanis \cite{darmanis2015survey} & Human & 420 & 22084 & 8  & 300\\
				
				GSE75748 cell & Chu \cite{chu2016single} & Human & 1018 & 19097 & 7 & 250\\
				
				GSE75748 time & Chu \cite{chu2016single} & Human & 758 & 19189 & 6 & 300 \\
				
				GSE82187 & Gokce \cite{gokce2016cellular} & Mouse & 685 & 18840 & 8 & 150 \\
				
				GSE84133 human (H) & Baron\cite{baron2016single} & Human & 1275 & 20125 & 6 & 150\\
				
				GSE84133 mouse (M) & Baron\cite{baron2016single} & Mouse & 782 & 14878 & 6 & 250 \\
				
				GSE94820 & Villani \cite{villani2017single} & Human & 1140 & 26593 & 5 & 250 \\ \hline
			\end{tabular}
			\label{tab: dataset}
		\end{table}

Since UMAP and t-SNE requires sufficient numbers of neighbors to perform visualization, for each data, cell types with less than 15 samples were removed. Then, log-transform was applied. The number of samples shown in \autoref{tab: dataset} is the filtered data.

For CCP, exponential kernel with $\tau = 6.0$ and $\kappa = 2.0$ was used. For UMAP, the default parameters with n$\_$neighbors = 15 was used used. For t-SNE, the default parameters from the sklearn's package with perplexity = 30 was used. In addition, PCA initialization was used.

\subsection{Visualization }
\begin{figure}[H]
	\centering
	\includegraphics{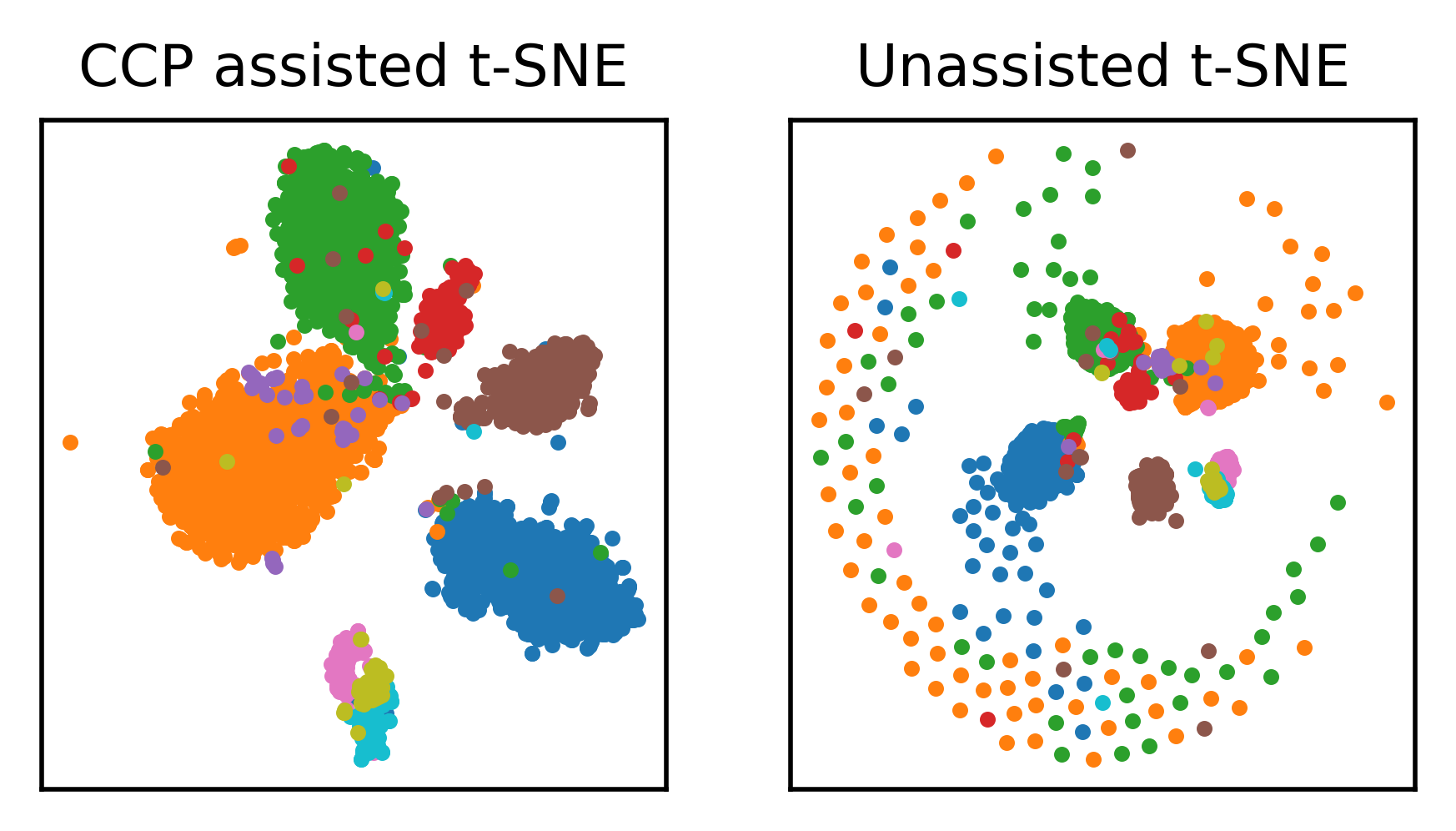}
	\caption{Comparison of visualization effects between  CCP-assisted t-SNE and unassisted t-SNE for the Baron dataset \cite{baron2016single}.  The samples were colored according to 9  cell types provided by the original authors. }
	\label{fig: bad tsne}
\end{figure}

Preprocessing of scRNA-seq data is a key step for visualization. \autoref{fig: bad tsne} shows an example of CCP-assisted t-SNE visualization and the original t-SNE visualization of the Baron dataset \cite{baron2016single}. The original data has 20,125 genes, and aggressively reducing the original dimension to 2 dimensions by t-SNE leads to poor visualization. In CCP-assisted t-SNE, CCP was utilized to reduce the original genes into 200 super-genes, which were further reduced to 2 dimensions with t-SNE for visualization. Obviously, CCP-assisted t-SNE significantly improves the visualization quality in this case. We further showcase CCP-assisted visualization on 8 more datasets shown in \autoref{tab: dataset}.

\begin{figure}[H]
\centering
	\includegraphics[width = \textwidth]{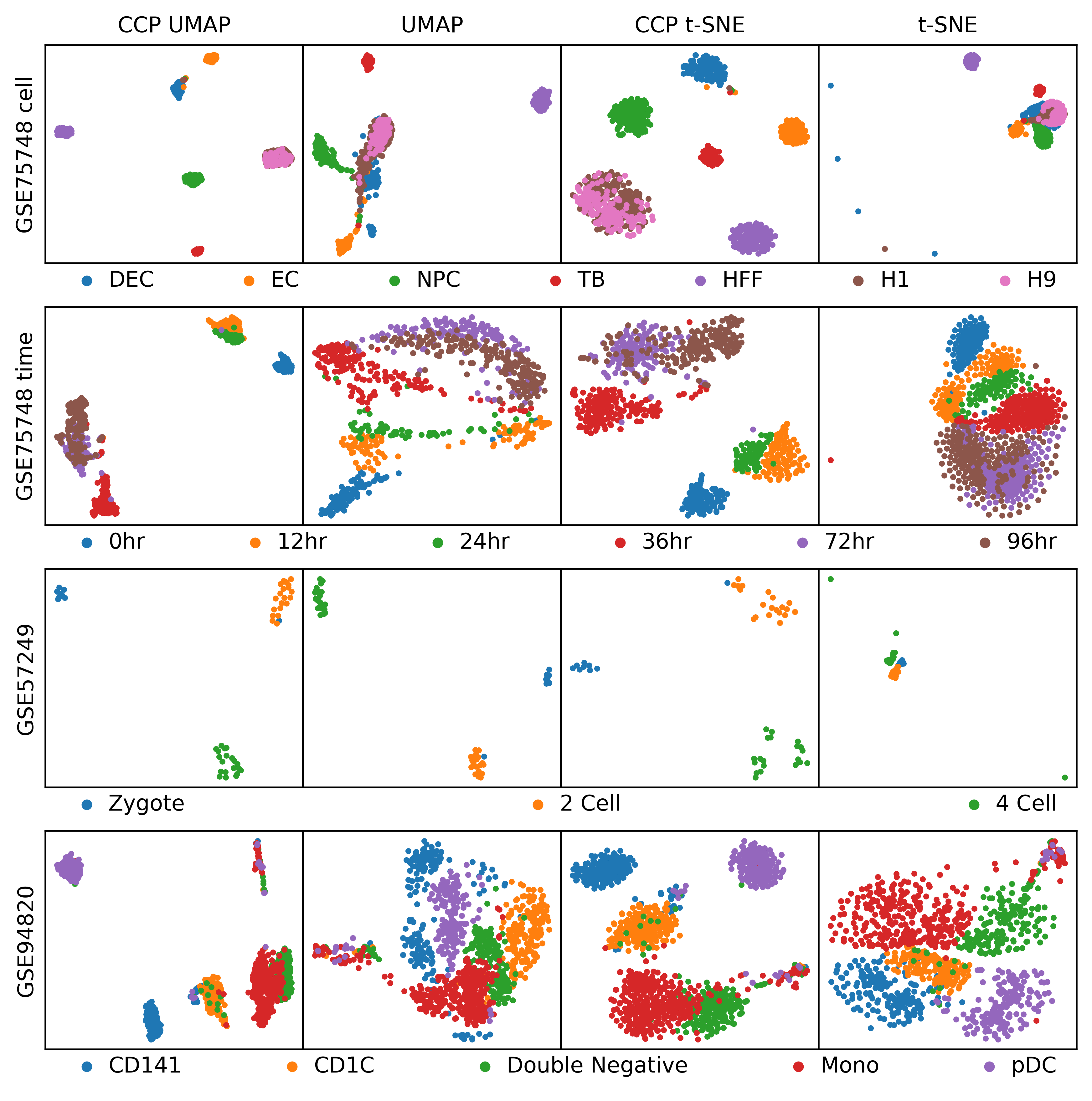}
	\caption{Comparison of visualization effects among  CCP-assisted UMAP and CCP-assisted  t-SNE,  standard UMAP, and standard t-SNE. Rows from top to bottom correspond to GSE75748cell, GSE75748time, GSE54279, and GSE94820 datasets. Columns from left to right correspond to CCP-assisted UMAP, standard UMAP, CCP-assisted t-SNE, and standard t-SNE. Before applying dimensionality reduction, the raw scRNA-seq data was log-transformed. For CCP-assisted visualization, the exponential kernel with $\tau = 6.0$ and $\kappa = 2.0$ was used. The sample entries were colored according to their true cell types provided by the original authors.}
	\label{fig: results1}
\end{figure}

\autoref{fig: results1} shows the comparison between CCP-assisted UMAP and t-SNE and the unassisted UMAP and t-SNE visualizations. The rows from top to bottom correspond to GSE75748cell, GSE75748time, GSE57249, and GSE94820 datasets, respectively. The columns from left to right correspond to CCP-assisted UMAP, unassisted UMAP, CCP-assisted t-SNE, and unassisted t-SNE, respectively. For the CCP-assisted visualization, CCP was applied using the exponential kernel with $\tau = 6.0$ and $\kappa = 2.0$. The sample entries were colored according to their true cell types.

Overall, CCP's super-genes are effective in improving UMAP and t-SNE visualization. The following are the key findings of the result.

\paragraph*{GSE75748 cell}
In all the visualizations of GSE75748 cell data, the undifferentiated H1 and H9 samples were relatively homogeneous, which is consistent with the findings by Chu et al. \cite{chu2016single}. However, in CCP-assisted UMAP, endothelial cells (EC, mesoderm derivatives), neuronal progenitor cells (NPC, ectoderm derivatives), and definitive endoderm cells (DEC, endoderm derivatives) samples are clearly distinct from the undifferentiated H1 and H9 samples. In the unassisted t-SNE visualization, all samples, besides the human foreskin fibroblasts (HFF) sample, are clustered together, whereas in CCP-assisted t-SNE, the samples are well-clustered.

\paragraph*{GSE75748 time}

In CCP-assisted visualization of GSE75748 time data, Chu et al. \cite{chu2016single} sequenced human embryonic stem cells at different time points as they differentiated in a hypoxic environment. Notice that in all the visualizations, samples sequenced at 72 hours and 96 hours are clustered together. This is consistent with Chu's finding that the cells most likely have completed their differentiation by the 72-hour mark. Additionally, notice that in CCP-assisted UMAP and CCP-assisted t-SNE, the original samples (0 hours) form a distinct cluster and are separated from the other clusters. However, the unassisted UMAP and unassisted t-SNE do not provide a clear separation. The 12-hour and 24-hour samples form a supercluster with a clear boundary in CCP-assisted UMAP and CCP-assisted t-SNE, whereas the unassisted counterparts do not show distinct clusters nor a supercluster.

\paragraph*{GSE57249}
CCP-assisted UMAP and unassisted UMAP yield similar results, where one zygote sample is clustered with the 2-cell stage cluster. In unassisted t-SNE, the 4-cell stage samples are not clustered together, and the zygote and 2-cell stage samples form one large cluster. However, in CCP-assisted t-SNE, the samples are properly clustered, and similar to UMAP, one zygote sample is clustered with the 2-cell stage samples.

\paragraph*{GSE94820}

The CCP-assisted visualization of GSE94820 shows a much clearer separation among the cell clusters compared to their unassisted counterparts. In particular, the unassisted UMAP and t-SNE do not exhibit clear separation among the cell types, which can be problematic for downstream analysis as different initializations can lead to different results. However, CCP-assisted UMAP and t-SNE indicate four main clusters: plasmacytoid dendritic cells (pDC), CD141$^+$ (CD141), CD1C1$^+$ (CD1C1), and a supercluster consisting of monocytes and CD141$^-$CD141$^-$ (Double Negative) cells. The clustering of monocytes and the Double Negative supercluster is consistent with the findings by Villani et al. \cite{villani2017single}, where they discovered shared key marker genes between these two cell types. Furthermore, Villani et al. also observed similar gene expression in a subset of CD1C1 cells, which may have caused some monocytes and Double Negative samples to be clustered with the CD1C1 cluster.

\begin{figure}[H]
	\centering
	\includegraphics[width = \textwidth]{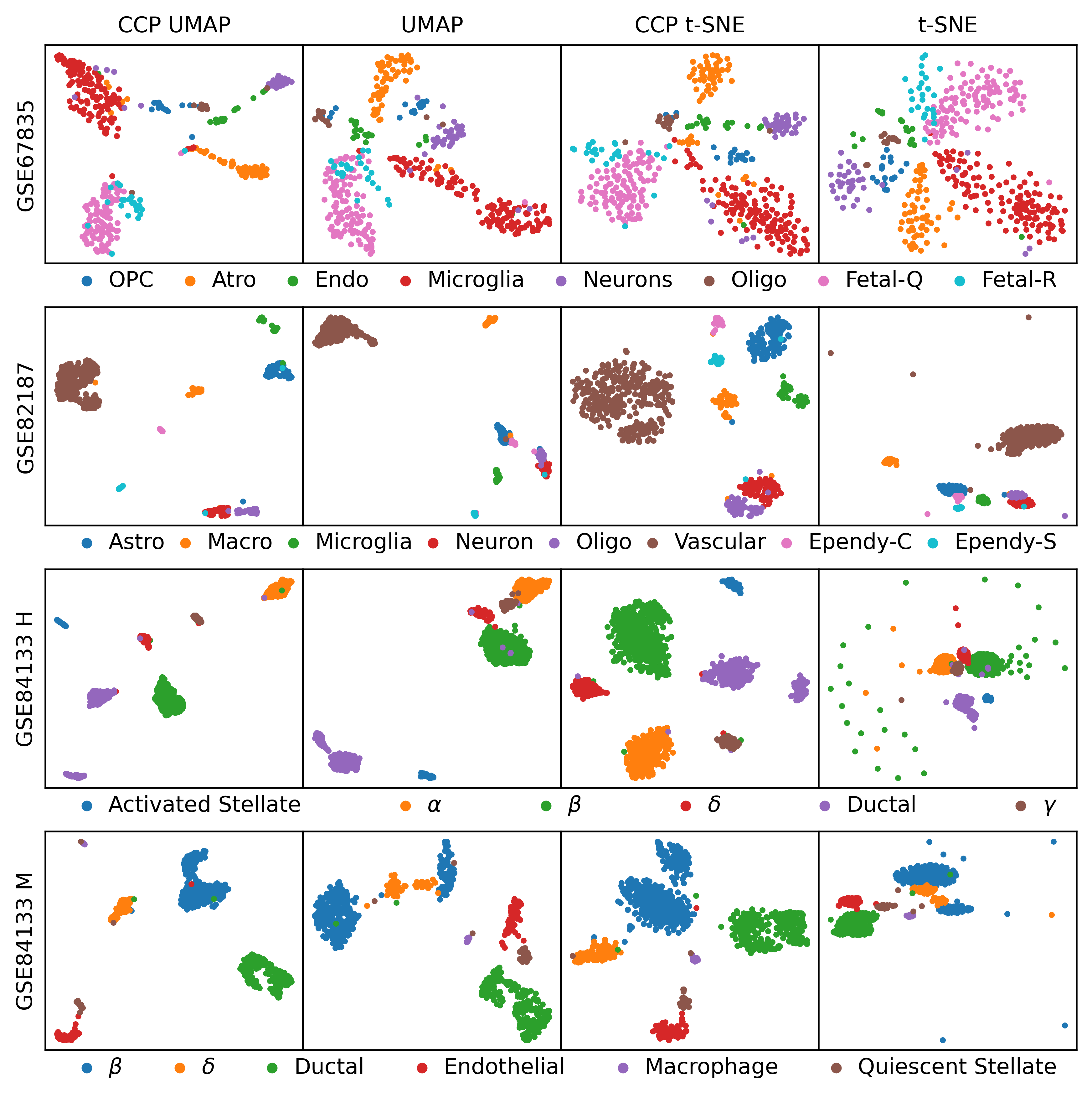}
	\caption{Comparison of CCP-assisted UMAP and t-SNE, and standard UMAP and t-SNE visualization. Rows correspond  to GSE67835, GS82187, GSE84133 human (H), GSE84133 mouse (M) data.   Columns correspond  to CCP-assisted UMAP, standard UMAP, CCP-assisted t-SNE and standard t-SNE visualization. Before applying dimensionality reduction, the raw scRNA-seq data was log transformed. For CCP-assisted visualization, exponential kernel with $\tau = 6.0$ and $\kappa = 2.0$ was used. The samples were colored according to their true cell types provided by the original authors.}
	\label{fig: results2}
\end{figure}

\autoref{fig: results2} shows the comparison between CCP-assisted UMAP and t-SNE and the unassisted UMAP and t-SNE visualizations. Each row, from top to bottom, corresponds to GSE67835, GSE82187, GSE84133 human, and GSE84133 mouse data, respectively. The columns, from left to right, correspond to CCP-assisted UMAP, unassisted UMAP, CCP-assisted t-SNE, and unassisted t-SNE, respectively. The raw data was log-transformed. For the assisted visualization, CCP was applied using the exponential kernel with $\tau = 6.0$ and $\kappa = 2.0$. The samples were colored according to their true cell types.

Overall, CCP's super-gene  approach is comparable to the unassisted visualization. The following are the key findings from the visualization.

\paragraph*{GSE67835}
The CCP-assisted and unassisted counterparts are quite similar in the GSE67835 data. In the CCP-assisted visualization, oligodendrocytes form a more compact cluster, and there is a clear distinction between neurons and the other cell clusters.

\paragraph*{GSE82187}
In GSE82187, the CCP-assisted visualization is able to clearly differentiate astrocytes from ependy-c samples, whereas the unassisted counterpart shows one cluster containing both samples.

\paragraph*{GSE84133 human}

CCP-assisted visualization of GSE84133 human data shows a clear separation in the clusters of beta and gamma. However, the beta samples are not as clustered in the CCP-assisted visualization. There are clearly two subclusters of ductal cells in all the visualizations, suggesting the presence of two subtypes within the ductal cells.
\paragraph*{GSE84133 mouse}

The CCP-assisted visualization of GSE84133 mouse data is comparable to the unassisted visualizations. However, it is worth noting that the unassisted visualization displays two subclusters of $\delta$ cells, whereas the CCP-assisted visualization shows only one cluster of $\delta$ cells.

\subsection{Accuracy}

\begin{figure}[H]
	\centering
	\includegraphics{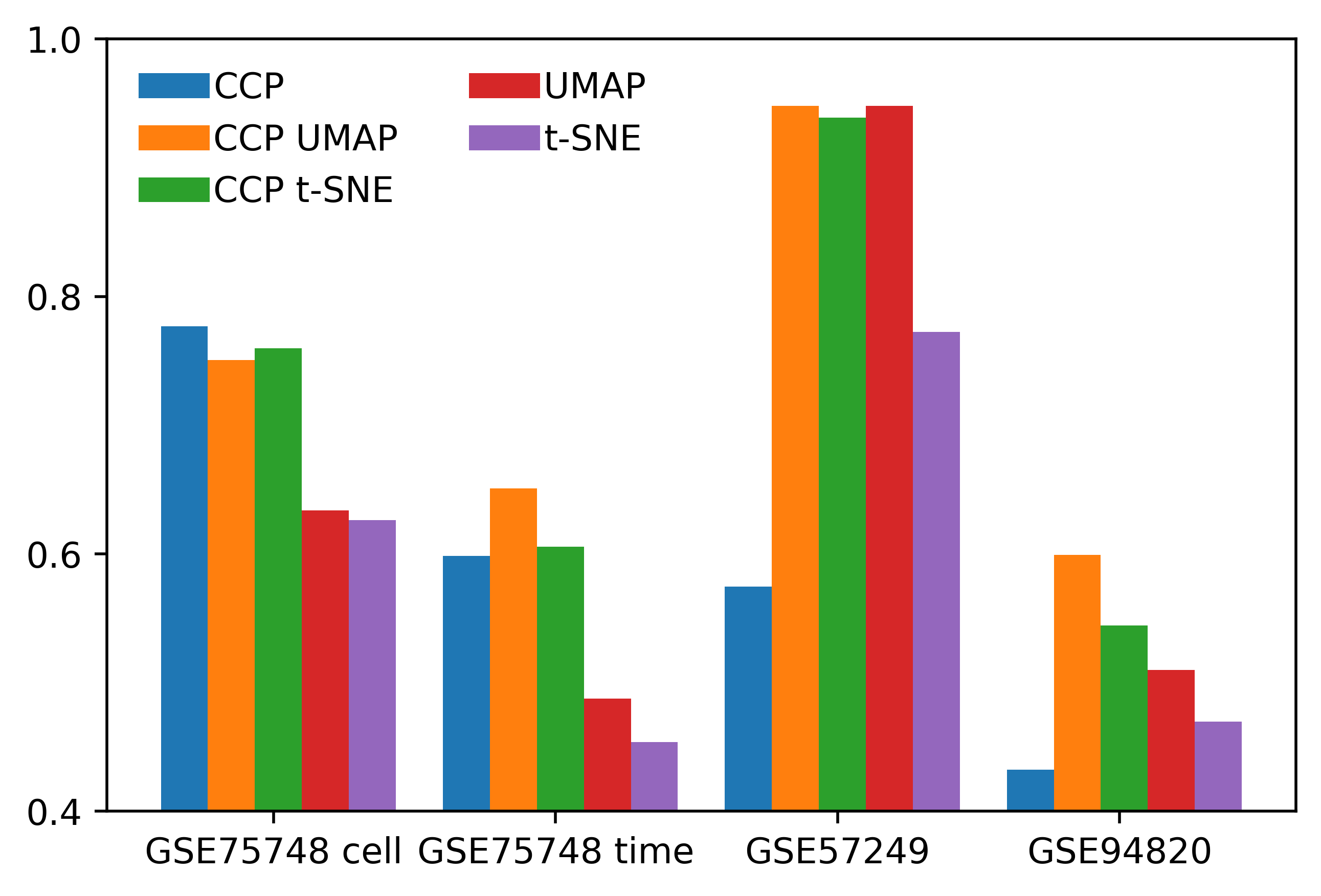}
	\caption{Comparison of five approaches for the ARI values of GSE75748 cell, GSE75748 time, GSE57249, and GSE94820 after dimensionality reduction. Blue bars correspond  to the clustering results of CCP. Orange and green lines correspond  to applying a secondary reduction of UMAP and t-SNE on the CCP super-genes. The red and purple bars correspond to unassisted UMAP and unassisted t-SNE, respectively.}
	\label{fig: result1 ari}
\end{figure}

\autoref{fig: result1 ari} shows the comparison between CCP, CCP-assisted UMAP and t-SNE, and the unassisted UMAP and t-SNE clustering results for GSE75748 cell, GSE75748 time, GSE57249, and GSE94820. For each dataset, 10 random seeds were used to perform the dimensionality reduction, and 30 random seeds were used to obtain the clustering results. The average Adjusted Rand Index (ARI) \cite{rand1971objective} is presented in \autoref{fig: result1 ari}.

For GSE75748 cell and GSE75748 time, CCP-assisted UMAP and t-SNE outperform their unassisted counterparts. In the case of GSE57249, CCP-assisted UMAP and the unassisted CCP yield similar results. However, the unassisted t-SNE performs significantly worse compared to the CCP-assisted t-SNE. In the case of GSE94820, although CCP has lower performance compared to other dimensionality reduction methods, CCP-assisted UMAP and t-SNE achieve higher ARI compared to the unassisted counterparts. This improvement in ARI may be attributed to the clustering nature in the UMAP and t-SNE algorithms.

\begin{figure}[H]
	\centering
	\includegraphics{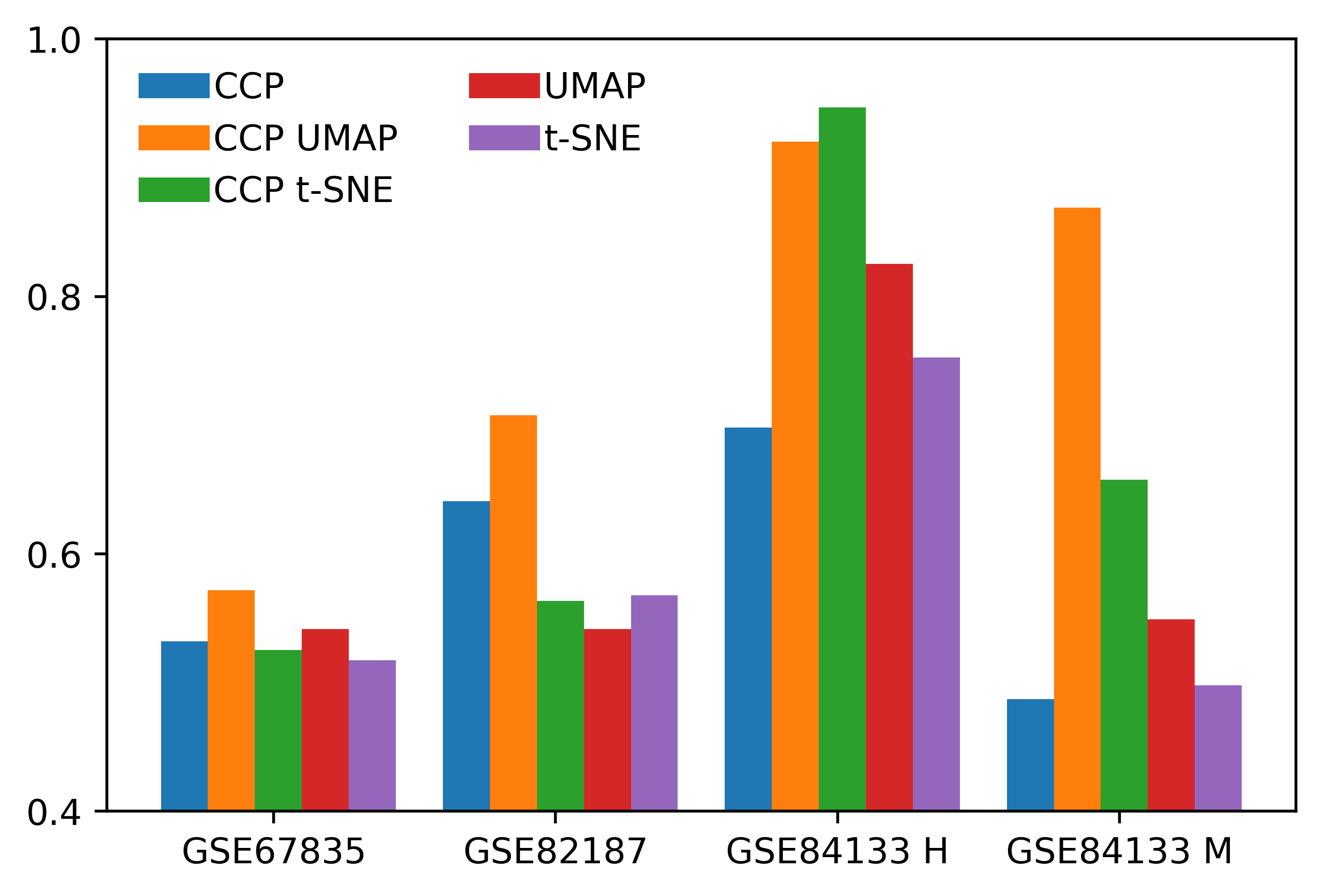}
	\caption{Comparison of five approaches for the ARI values of GSE75748 cell, GSE75748 time, GSE84133 h3, and GSE94820 after dimensionality reduction. Blue bars correspond  to the clustering results of CCP. Orange and green lines correspond  to applying a secondary reduction of UMAP and t-SNE on the CCP super-gene. The red and purple bars correspond to unassisted UMAP and t-SNE, respectively.}
	\label{fig: result2 ari}
\end{figure}

\autoref{fig: result2 ari} illustrates the comparison among CCP, CCP-assisted UMAP, CCP-assisted t-SNE, unassisted UMAP, and unassisted t-SNE clustering results for GSE67835, GSE82187, GSE84133 h4, and GSE84133 m1. In each dataset, 10 random seeds were used for dimensionality reduction, and 30 random seeds were used to obtain the clustering results. The average Adjusted Rand Index (ARI) is presented in \autoref{fig: result2 ari}.

For GSE67835, all clustering results are comparable, with CCP-assisted UMAP and CCP-assisted t-SNE slightly outperforming their unassisted counterparts. In the case of GSE82187, CCP-assisted UMAP exhibits significantly better performance than the unassisted UMAP, while CCP-assisted t-SNE performs similarly to the unassisted t-SNE. For GSE84133 h4 and GSE84133 m1, both CCP-assisted UMAP and CCP-assisted t-SNE outperform the unassisted counterparts.

\section{Discussion}

\subsection{Mean accuracy}

\begin{figure}[H]
	\centering
	\includegraphics{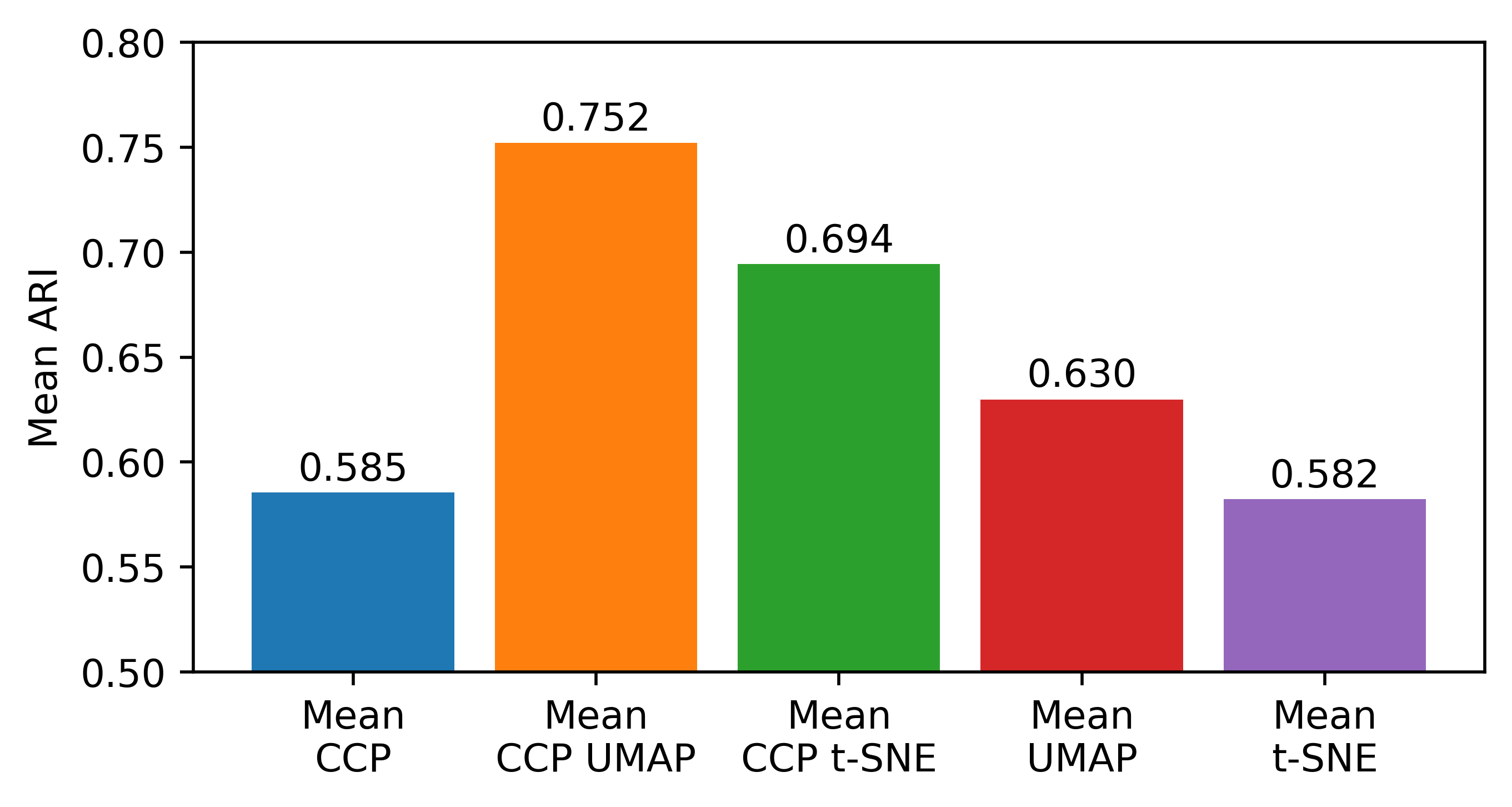}
	\caption{Mean ARI of CCP, CCP-assisted UMAP, CCP-assisted t-SNE,    UMAP, and   t-SNE across the 8 dataset described in \autoref{tab: dataset}.}
	\label{fig: mean ari}
\end{figure}

\autoref{fig: mean ari} shows the average  ARI over datasets of CCP,  CCP-assisted UMAP, CCP-assisted  t-SNE,  UMAP, and t-SNE. It  is clear that CCP initialization improves both UMAP  and t-SNE performance. On average, CCP improves UMAP's  and t-SNE's performance by 19\%. These improvements are very significant. 

\subsection{Effect of LV-gene}
We have utilized the LV-gene to enhance the predictive power of super-genes with low variance. By using a high cutoff ratio, we can reduce the number of genes used in the feature partition, potentially resulting in a lower number of super-genes. To assess the impact of the cutoff ratio on the number of super-genes used for UMAP and t-SNE visualization, we conducted tests.

\begin{figure}[H]
	\centering
	\includegraphics{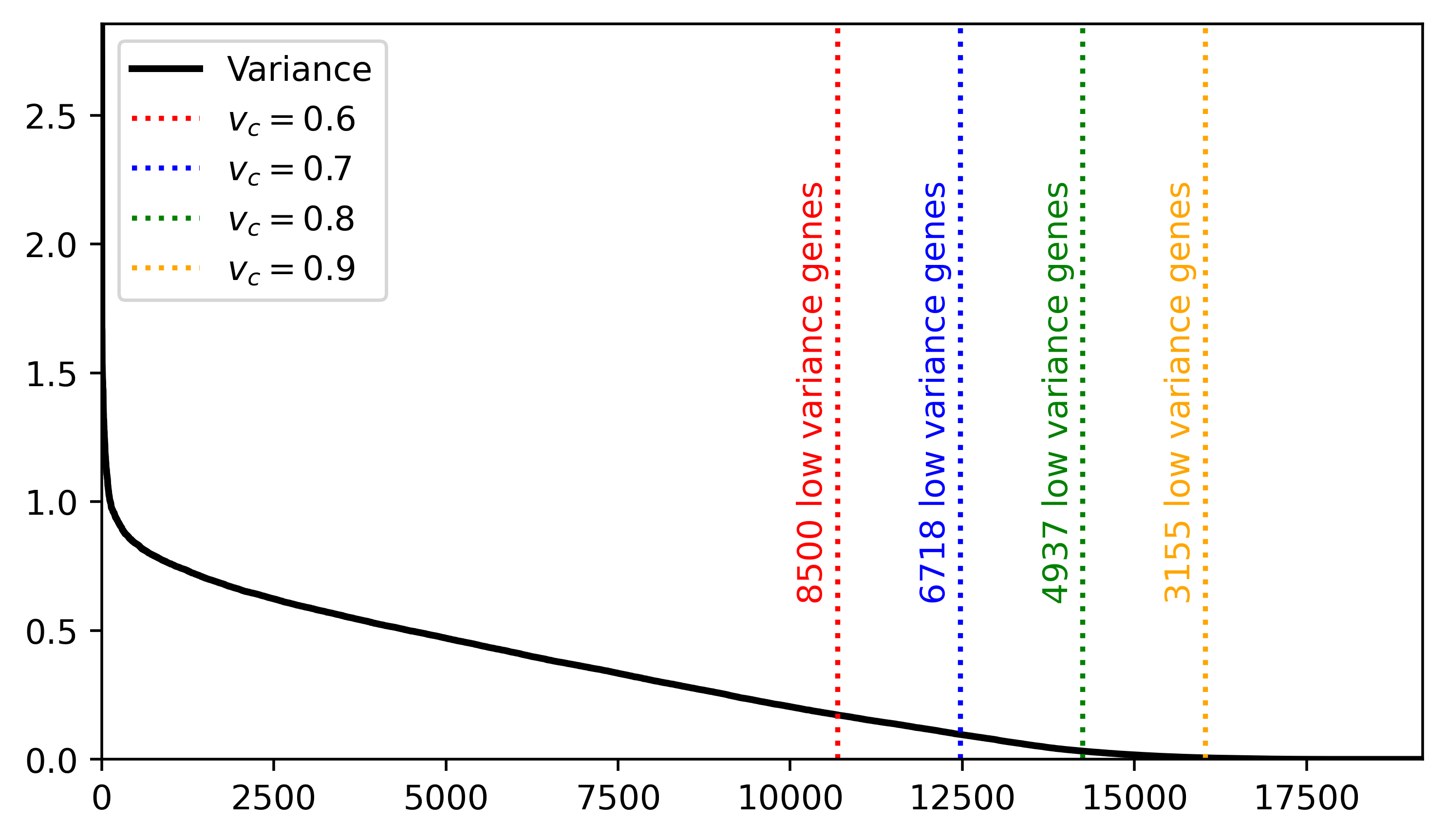}
	\caption{Variance of each genes and the variance cutoff for the LV-gene. The black line shows the variance in descending order. Red, blue, green, and orange vertical line shows the variance cutoff ratio of 0.6, 0.7, 0.8, and 0.9, respectively. The text shows the number of genes that are grouped into a LV-gene.}
	\label{fig: GSE75748time variance}
\end{figure}

\autoref{fig: GSE75748time variance} presents the variance of the genes in GSE75748 time data with respect to the variance cutoff ratio for the LV-gene. The black line represents the variance in descending order. The red, blue, green, and orange dashed lines indicate the variance cutoff ratios ($v_c$) at 0.6, 0.7, 0.8, and 0.9, respectively. The LV-gene selects all the genes above the cutoff threshold.

.

Notably, when $v_c = 0.8$ and 0.9, despite containing 4937 and 3155 genes, respectively, all the genes exhibit low variance. At $v_c = 0.7$, the LV-gene still includes low variance genes, but also incorporates more genes with potential predictive power that may not be captured using traditional approaches

\begin{figure}[H]
	\includegraphics[width = \textwidth]{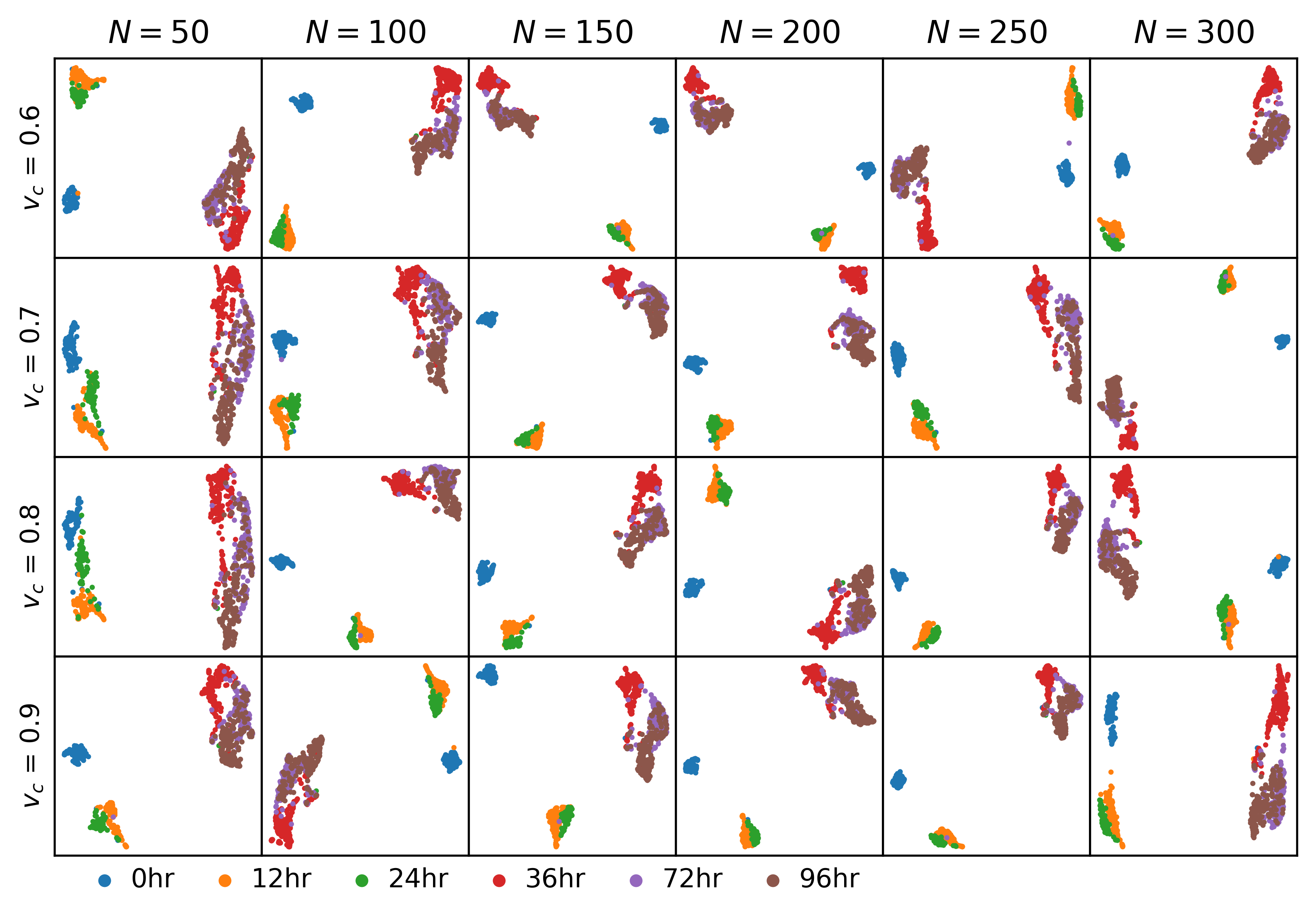}
	\caption{Comparison of the variance cutoff $v_c$ and the number of super-genes used in CCP on the UMAP visualization. The rows correspond to the varying variance cutoff ratio, and the columns corresponds to the different number of super-genes used. Exponential kernel with $\tau = 6.0$ and $\kappa = 2.0$ was used for CCP.
	}
	\label{fig: GSE75748 cutoff umap}
\end{figure}

\autoref{fig: GSE75748 cutoff umap} illustrates the impact of variance cutoff and the number of super-genes used in the CCP initialization on UMAP visualization for GSE75748 time data. The rows, from top to bottom, correspond to the variance cutoff values $v_c = 0.6, 0.7, 0.8, $ and $ 0.9$, while the columns, from left to right, represent different numbers of super-genes $N = 50, 100, 150, 200, 250,$ and $300$.

Noticeably, when the number of super-genes is set to $N = 50$ and $100$, the samples at 36 hours are clustered together with the clusters at 72-96 hours for all variance cutoff values. However, as the number of super-genes increases to $N = 150-300$, a clear distinction emerges between the two clusters for variance cutoff values of $v_c = 0.7$ and $0.8$. Conversely, when the variance cutoff is set to $v_c = 0.9$, the 36-hour samples do not form a distinct cluster, suggesting that the super-genes may contain noise, making it challenging to differentiate the 36-hour samples from the other types.

\begin{figure}[H]
	\includegraphics[width = \textwidth]{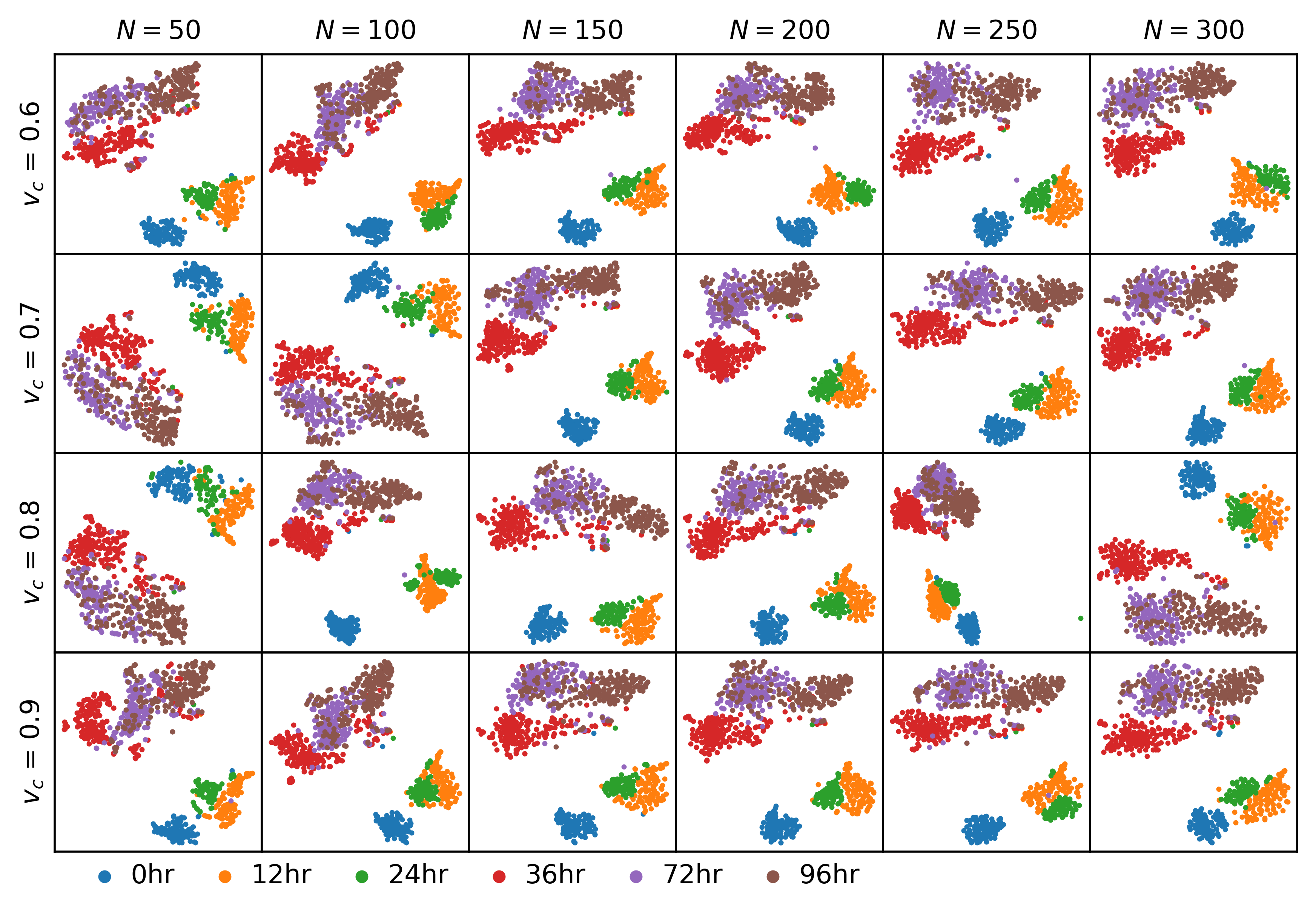}
	\caption{Impacts  of the variance cutoff $v_c$ and the number of super-genes used in CCP on the t-SNE visualization. The rows correspond to the varying variance cutoff ratio, and the columns correspond  to the different numbers of super-genes used. Exponential kernel with $\tau = 6.0$ and $\kappa = 2.0$ was used for CCP.
	}
	\label{fig: GSE75748 cutoff tsne}
\end{figure}

\autoref{fig: GSE75748 cutoff tsne} illustrates the impact of variance cutoff and the number of super-genes used in the CCP initialization on t-SNE visualization for GSE75748 time data. The rows from top to bottom correspond to the variance cutoff values $v_c = 0.6, 0.7, 0.8,$ and $0.9$, while the columns from left to right correspond to different numbers of super-genes $N = 50, 100, 150, 200, 250,$ and $300$.

For all variance cutoff values, when $N = 50$ and $100$, the CCP-assisted t-SNE shows a large super cluster consisting of 36hrs, 72hrs, and 96hrs samples. However, when $N = 300$ for $v_c > 0.6$, the 36hrs samples form their own cluster. This suggests that a variance cutoff of $v_c = 0.6$ may have filtered out too many useful genes. In Chu et al. \cite{chu2016single}, the authors suggested that although samples were sequenced at precise times, individual cells may be in different states of their cell cycle, leading to potential variability within the samples within the same clusters. This suggests that the genes grouped in the LV-gene may have small variation between the different clusters and should not be ignored or dropped.

\section{Conclusion}
CCP is a nonlinear data-domain dimensionality reduction technique that leverages gene-gene correlations to partition genes and utilizes cell-cell correlations to generate super-genes. Unlike methods that involve matrix diagonalization, CCP can be directly applied as a primary dimensionality reduction tool to complement traditional visualization techniques like UMAP and t-SNE. In our experiments with eight datasets, CCP-assisted UMAP and CCP-assisted t-SNE visualizations consistently outperformed or were comparable to the unassisted versions. Additionally, CCP's super-genes have the ability to capture continuous changes during cell differentiation, which is advantageous for visualization purposes. Furthermore, we demonstrated that CCP-assisted UMAP and CCP-assisted t-SNE  improve the accuracy of the original UMAP and t-SNE algorithms by  19\% in each case.

	\section{Code and Data availability}
	All data can be downloaded from Gene Expression Omnibus \cite{edgar2002gene, barrett2012ncbi}. The processing files for these data can be found on \hyperlink{https://github.com/hozumiyu/SingleCellDataProcess}{https://github.com/hozumiyu/SingleCellDataProcess}. CCP is made available through our web-server at \hyperlink{https://weilab.math.msu.edu/CCP/}{https://weilab.math.msu.edu/CCP/} or through the source code \hyperlink{https://github.com/hozumiyu/CCP}{https://github.com/hozumiyu/CCP} . The code to reproduce this paper is found at \hyperlink{https://github.com/hozumiyu/CCP-scRNAseq-UMAP-TSNE}{https://github.com/hozumiyu/CCP-scRNAseq-UMAP-TSNE}

	\section{Acknowledgments}
This work was supported in part by NIH grants  R01GM126189,  R01AI164266, and R35GM148196
, NSF grants DMS-2052983,  DMS-1761320, and IIS-1900473,  NASA grant 80NSSC21M0023,  MSU Foundation,  Bristol-Myers Squibb 65109, and Pfizer.

 \end{document}